\documentclass{article}
\usepackage{spconf,amsmath,epsfig}
\usepackage{tabularx}
\usepackage{ltablex}
\usepackage{graphicx}
\usepackage{todonotes}
\keepXColumns

\let\OLDthebibliography\thebibliography
\renewcommand\thebibliography[1]{
  \OLDthebibliography{#1}
  \setlength{\parskip}{0pt}
  \setlength{\itemsep}{0pt plus 0.3ex}
}

\usepackage{adjustbox} 
\newcommand{\winoground}{{\fontfamily{lmtt}\selectfont Winoground}}

\newcommand{\raven}{{\fontfamily{lmtt}\selectfont Raven-50}}

\newcommand{\factify}{{\fontfamily{lmtt}\selectfont Factify2}}

\newcommand{\factifyv}{{\fontfamily{lmtt}\selectfont Factify-V}}

\begin{document}\sloppy

% Example definitions.
% --------------------
\def\x{{\mathbf x}}
\def\L{{\cal L}}

% Title.
% ------
\title{CoCoT: Contrastive Chain-of-Thought Prompting  for Large Multimodal Models with Multiple Image Inputs}
%
% Single address.
% ---------------

\name{Daoan~Zhang$^{1, *}$, Junming~Yang$^{2, *}$, Hanjia~Lyu$^{1, *}$, Zijian~Jin$^{3}$, Yuan~Yao$^{1}$, Mingkai~Chen$^{4}$, Jiebo~Luo$^{1}$}

\address{$^1$ University of Rochester, $^2$ Nanjing University of Posts and Telecommunications, \\ $^3$ New York University, $^4$ Stony Brook University, \\
$^*$ These authors contributed equally to this work.}

\maketitle

\begin{abstract}
When exploring the development of Artificial General Intelligence (AGI), a critical task for these models involves interpreting and processing information from multiple image inputs. However, Large Multimodal Models (LMMs) encounter two issues in such scenarios: (1) a lack of fine-grained perception, and (2) a tendency to blend information across multiple images. We first extensively investigate the capability of LMMs to perceive fine-grained visual details when dealing with multiple input images. The research focuses on two aspects: first, image-to-image matching (to evaluate whether LMMs can effectively reason and pair relevant images), and second, multi-image-to-text matching (to assess whether LMMs can accurately capture and summarize detailed image information). We conduct evaluations on a range of both open-source and closed-source large models, including {\sc GPT-4V}, {\sc Gemini}, {\sc OpenFlamingo}, and {\sc MMICL}. To enhance model performance, we further develop a \underline{Co}ntrastive \underline{C}hain-\underline{o}f-\underline{T}hought (CoCoT) prompting approach based on multi-input multimodal models. This method requires LMMs to compare the similarities and differences among multiple image inputs, and then guide the models to answer detailed questions about multi-image inputs based on the identified similarities and differences. Our experimental results showcase CoCoT's proficiency in enhancing the multi-image comprehension capabilities of large multimodal models.
\end{abstract}
\begin{keywords}
Large Multimodal Model, Multimodal Prompting, Large Language Model 
\end{keywords}
\section{Introduction}
\label{sec:intro}

Recent advancements in Large Language Models (LLMs)~\cite{touvron2023llama, ouyang2022training, zhang2023dnagpt} have sparked optimism in the pursuit of Artificial General Intelligence (AGI). Given the pivotal role of vision in human information acquisition, its integration is crucial for AGI's perceptual capabilities. To bridge the gap between textual and visual modalities, researchers are experimenting with aligning language with vision~\cite{liu2023visual, tang2023llmva} and directly encoding visual inputs into discrete tokens~\cite{hu2022promptcap, hua2022fine}. These efforts have demonstrated the substantial potential of large multimodal models in processing multimodal content. However, they still fall short of human-like perception of the world~\cite{lyu2023gpt, yu2022unbiased}. One significant challenge is the \textit{loss of image detail} when using natural language, a medium less precise than visual data. Complex visual information, such as subtle lighting shifts or intricate patterns, often requires comprehensive verbal description. Another hurdle is understanding the \textit{relationship between multiple image inputs}. Language-based descriptions of relationships and interactions within and across images can become challenging, necessitating explanations of both individual elements and their spatial and contextual ties. This complexity often results in ambiguous or overly extensive explanations, highlighting the limitations of current models in emulating human-like perception.
% (1) a loss of image details, and (2) difficulty in understanding the interaction across multiple image inputs.
To address these two issues, researchers have developed various multimodal prompting strategies~\cite{zheng2023ddcot, mitra2023compositional} in an attempt to guide LMMs in extracting essential information from the visual content for effective multi-image understanding. Although these methods have shown proficiency in comprehending single-image contexts, they encounter obstacles when it comes to discerning relationships between multiple images. This difficulty primarily stems from an \textit{insufficient focus on key information, which requires joint consideration of all images involved}.

\begin{figure*}[ht]
    \centering
    \includegraphics[width=.95\linewidth]{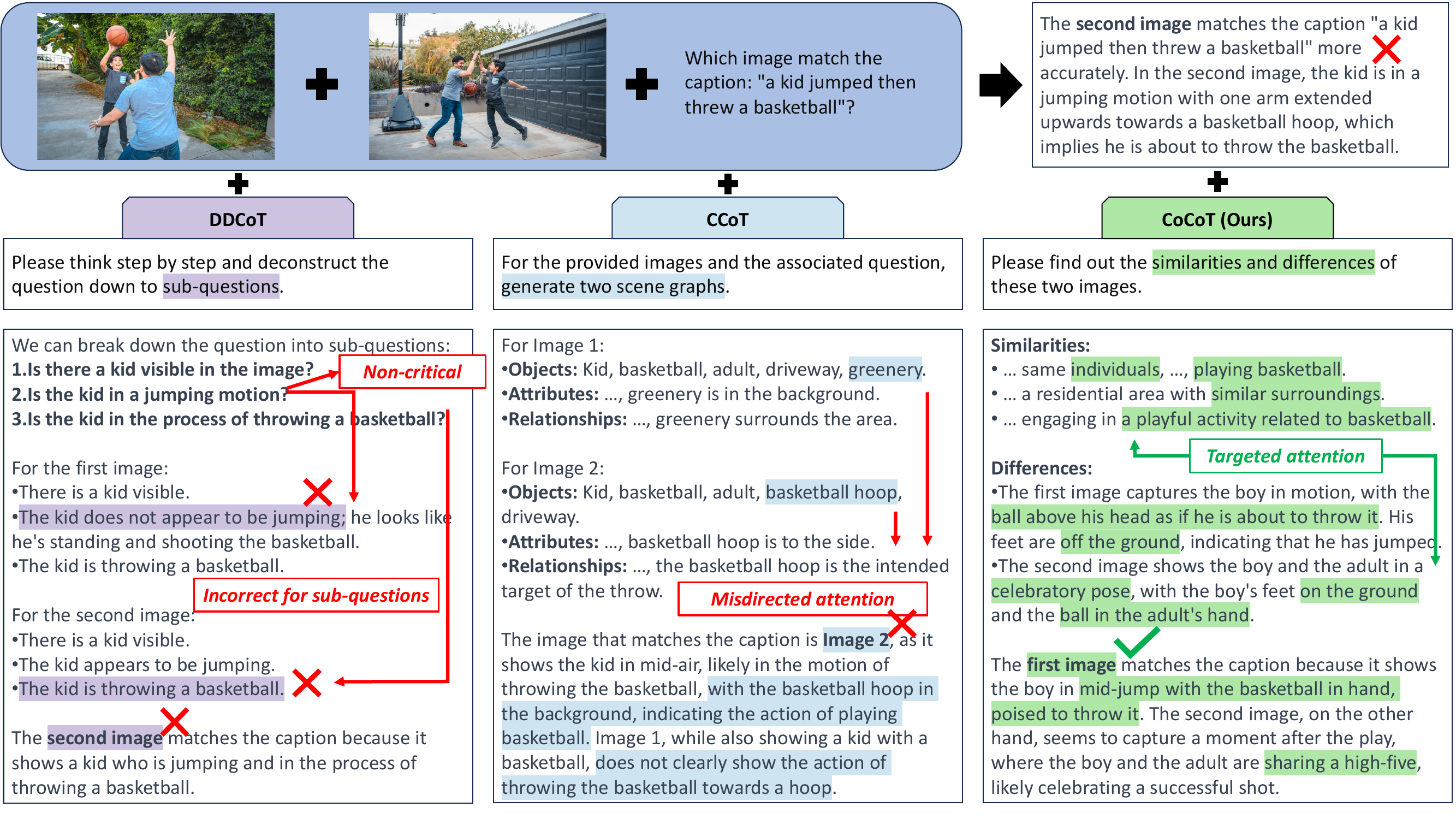}
    \vspace{-0.2in}
    \caption{Comparison between different multimodal prompting strategies. The unique components in each prompting strategy's corresponding response are highlighted in varied colors. Note that {\sc GPT-4V} is used in this example.
     }
             \vspace{-0.1in}
    \label{fig:enter-label}
\end{figure*}

In our study, we introduce \underline{Co}ntrastive \underline{C}hain-\underline{o}f-\underline{T}hought (CoCoT), a novel multimodal prompting strategy designed to overcome the limitations of existing prompting strategies and enhance LMMs' performance in multi-image tasks. CoCoT prompts LMMs to discern and articulate the \textbf{similarities and differences} among various inputs, laying the groundwork for answering detailed, multi-image-based questions (Fig.~\ref{fig:enter-label}). This method sharpens the models' focus, particularly on the distinctions between inputs, ensuring comprehensive capture of nuanced, question-relevant information during summarization. We rigorously evaluate CoCoT in two multi-image tasks. The first task, image-to-image matching, assesses the LMMs' competence in logically associating and pairing relevant images. The second task, image-to-text matching, tests the LMMs' ability to accurately capture and summarize detailed image data. Our evaluations span both widely-used open-source models and leading-edge closed-source models. %Through diverse testing settings, our study aims to demonstrate CoCoT's efficacy in advancing the multimodal understanding capabilities of LMMs, particularly in complex multi-image contexts.
To summarize, our main contributions are:
\begin{itemize}
    \item We find that most current models do \textit{not} perform well in fine-grained multi-image tasks.
    \item To address the issues with existing methods, we propose a novel \underline{Co}ntrastive \underline{C}hain-\underline{o}f-\underline{T}hought (CoCoT) prompting strategy to enhance models' understanding of the relationships between multiple image inputs.
    \item Our proposed method produces significant improvement for both open-source and closed-source models.
\end{itemize}

\section{Related Work}

\textbf{Large Multimodal Models.}
Inspired by the advancements of LLMs (\textit{e.g.}, {\sc LLaMA}~\cite{touvron2023llama}), LMMs offer a promising way towards AGI with multimodal information. These models blend the textual reasoning prowess of LLMs with the image and video comprehension of Vision-and-Language models. This fusion enables LMMs to handle complex tasks requiring both a profound understanding and expressive generation across various modalities.
Several open-source LMMs like {\sc LLaVA}~\cite{liu2023visual} have emerged, demonstrating competence in tasks such as image captioning and visual question-answering. However, their architectural limitations restrict their understanding and reasoning to a single image. Conversely, models like {\sc OpenFlamingo}~\cite{awadalla2023openflamingo}, and {\sc MMICL}~\cite{zhao2023mmicl} employ specialized architectures enabling the processing of multiple image features, which better mirrors real-world scenarios. Closed-source LMMs such as {\sc GPT-4V}~\cite{gpt4report} and {\sc Gemini}~\cite{team2023gemini} go beyond basic object descriptions to capture the scene's context~\cite{mitra2023compositional}, emotions~\cite{zhao2023mmicl}, and relationships~\cite{thrush2022winoground}.
A common technique to enhance performance is fine-tuning, but applying similar methods to LMMs presents computation challenges~\cite{liu2023mmbench}. To overcome this, we propose a novel approach to directly enable detailed analysis and reasoning on images without additional training data.
% Fine-tuning LMMs often requires vast amounts of multimodal data, which is costly and inconvenient, especially when dealing with multiple images.
%Contrastive learning techniques implemented in vision-language models \cite{radford2021learning} represent a significant leap forward in how computers learn to connect visual and textual representations. While effective for basic tasks, these methods struggle with complex downstream applications requiring text-image reasoning or generation, such as visual question-answering \cite{antol2015vqa,teney2018tips}. Leveraging LLMs' reasoning and vision-language representations, it allows LMMs to generate richer and more informative captions for images, going beyond basic object descriptions to capture the scene's context \cite{zhang2023prise}, emotions\cite{peng2023kosmos}, and relationships \cite{thrush2022winoground}.  

\begin{figure*}[t]
    \centering
    \includegraphics[width=.95\linewidth]{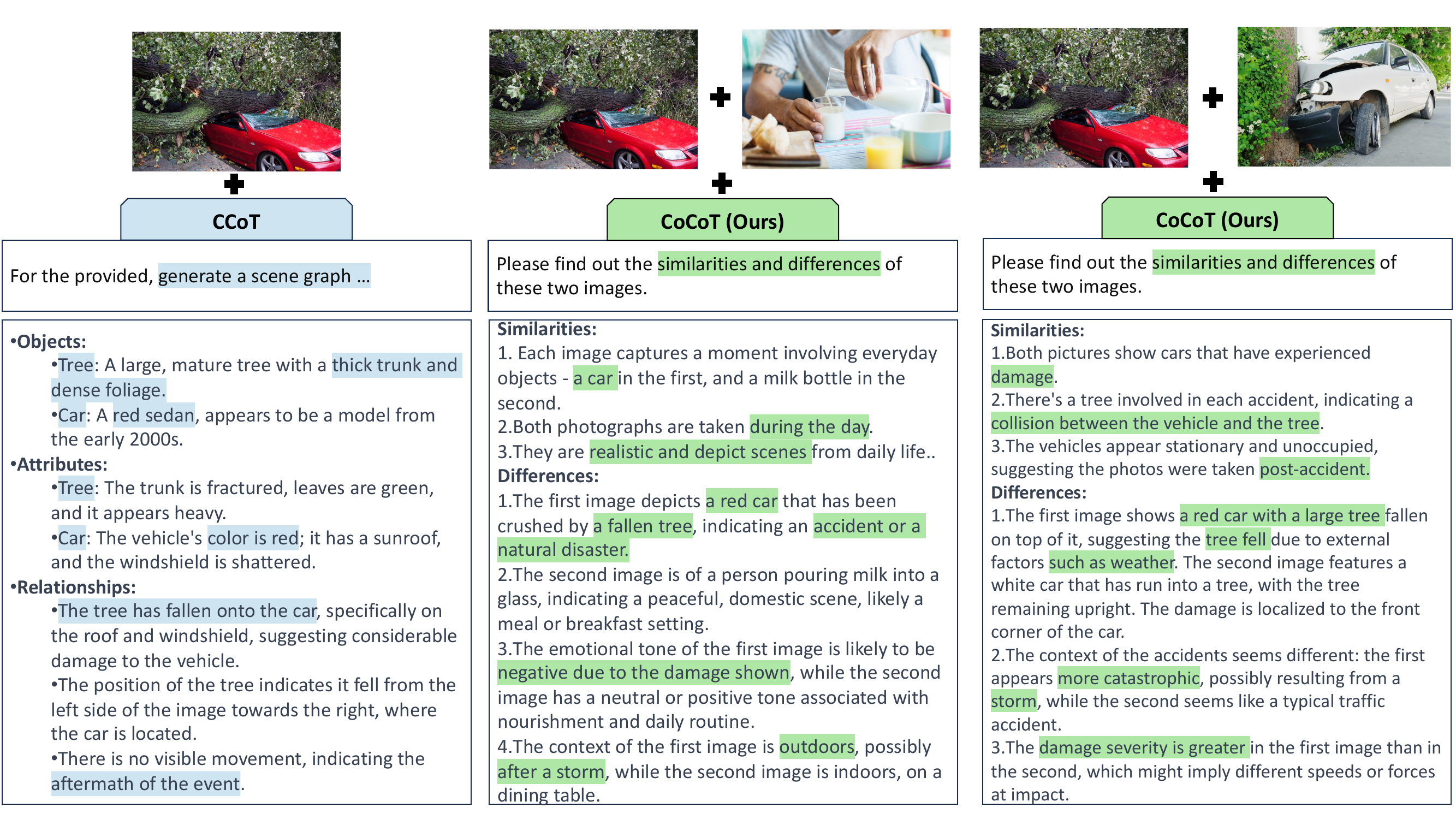}
      \vspace{-0.2in}
    \caption{Different CoT-based methods and their performance in extracting information from images under various conditions, with {\sc GPT-4V} being used in the experiments. Left: Utilizing CCoT to generate image information; Middle: CoCoT prompting between images with a big domain gap; Right: CoCoT prompting between images with a small domain gap.}
    \label{fig:compare}
          \vspace{-0.1in}
\end{figure*}

\noindent\textbf{Multimodal Prompting Methods.}
Within the domain of LLMs, several language prompt methods have been established to enhance inference capabilities and ensure accurate results during prediction. These include zero-shot~\cite{kojima2022large}, few-shot~\cite{zhang2023prompt}, and Chain-of-Thought (CoT)~\cite{zheng2023ddcot,mitra2023compositional} approaches. Recently, research has begun exploring the application of prompting techniques in the multimodal domain to improve the comprehension and reasoning abilities of LMMs for image data. Current multimodal prompts employed in LMMs often exhibit limitations in capturing the intricate interrelationships between visual and language information, particularly when faced with multi-image inputs. As shown in the example in Fig.~\ref{fig:enter-label}, they are not able to identify the critical action of the boy throwing the ball. To overcome this challenge, we propose a novel prompting method that directs LMMs to extract and analyze essential information, requiring a holistic consideration of all the input images.

%With LLMs and LMMs gaining traction, prompting methods like zero-shot \cite{kojima2022large}, few-shot \cite{zhang2023prompt}, expert prompting \cite{zamfirescu2023johnny}, and Chain-of-Thought (CoT) \cite{mitra2023compositional} with its extensions like self-consistency, Tree-of-Thought (ToT) \cite{long2023large}, and Graph-of-Thought (GoT) \cite{besta2023graph} have emerged as vital tools for harnessing their power. These techniques, implemented during inference time, enable precise control over model outputs and provide crucial context for their efficient utilization.

\section{Contrastive Chain-of-Thought}

% \begin{figure*}[t]
%     \centering
%     \includegraphics[width=0.8\textwidth]{main.png}
%     \caption{Full prompt example of CoCoT. The first step in our prompting method is to generate a comparison given both the image and textual task as context. Following this, the answer is extracted by prompting the LMM with the image, similarities and differences, questions. Prompt sections unique to our method are bolded.}
%     \label{fig:cocot_example}
% \end{figure*}

\subsection{Motivation Analysis}

Traditional CoT-based prompting methods for LMMs can be categorized into two types. The first type is based on text understanding, such as DDCoT (\textit{i.e.,} Duty-Distinct Chain-of-Thought)~\cite{zheng2023ddcot}, which decomposes a question into sub-questions for a step-by-step response. The second type is based on image understanding, like CCoT (\textit{i.e.,} Compositional Chain-of-Thought)~\cite{mitra2023compositional}, which generates a scene graph of the image to provide answers. However, while processing images, the text-based CoT does not enable LMMs to directly acquire and comprehend the detailed information in images. As shown in Fig.~\ref{fig:enter-label}, DDCoT does not enable the LMM to recognize that the kid in the second image is \textbf{not} throwing a basketball. The image-based CCoT merely extracts basic information about the main objects in the image, also overlooking significant details. As shown in Fig.~\ref{fig:enter-label}, CCoT generates a series of scene graphs unrelated to the question. Existing CoT-based prompting methods struggle to notice the details when answering questions about images rich in detail. Therefore, an effective prompting method should enable LMMs to discern and understand the details in images, and subsequently answer questions based on this understanding.

\subsection{Methodology}

We focus on how to enable LMMs to extract more detailed information from images, especially when the images are very similar. Initially, we examine the extent to which LMMs based on CCoT can extract information from images, as illustrated in Fig.~\ref{fig:compare}. {\sc GPT-4V}, utilizing CCoT, is limited to identifying entities, their characteristics, and straightforward details like events and relationships between entities. Drawing inspiration from contrastive learning, our approach encourages LMMs to discern similarities and differences within images. We discover that these models are capable of engaging with more complex information, such as reasoning, even when there is a considerable difference in the domain between the images being compared and the original. For instance, they might deduce that an image's scene likely follows a storm and recognize a negative emotional tone in it. When comparing similar images, focusing on the similarities and differences of images effectively highlights the contrasts, such as recognizing more severe damage in one image compared to another, or differentiating the causes of car damage between two images, thereby effectively facilitating causal reasoning. Consequently, we develope the Contrastive Chain-of-Thought prompting. As shown in Fig.~\ref{fig:enter-label}, this approach, similarly starting from an image perspective, initially compares the similarities and differences between various image inputs. It then directs LMMs to answer questions based on the insights gathered from such comparisons.

% \begin{figure}[t]
%     \centering
%     \includegraphics[width=0.25\textwidth]{cocot.png}
%     \caption{Contrastive Chain-of-Thought}
%     \label{fig:cocot}
% \end{figure}

% \subsection{CoCoT}
% \vspace{-0.2in}

\section{Experiments and Results}
\subsection{Experiment Setup}
\noindent\textbf{Datasets.} 
We evaluate the effectiveness of CoCoT on two fine-grained multi-image tasks: (1) image-to-image matching and (2) multi-image-to-text matching. Both tasks are well-suited for assessing whether the CoT-based method enables LMMs to acquire more fine-grained information from multiple image inputs. The image-to-image matching task employs the \raven~\cite{zhang2019raven, huang2023language} and \factify~\cite{suryavardan2023factify} datasets. This task tests the models' ability to identify and interpret visual details, requiring them to determine the degree of match between different images. For the multi-image-to-text matching task, we use \winoground~\cite{thrush2022winoground}. This task requires LMMs to effectively pair similar images with their corresponding textual descriptions, or alternatively, to align similar texts with the corresponding images. Details including dataset statistics and preprocessing specifics are discussed in Appendix~\ref{appendix_sec:dataset}.

\noindent\textbf{Baselines.} We compare CoCoT prompting to two state-of-the-art methods in CoT-based multimodal prompting. This includes DDCoT~\cite{zheng2023ddcot} and CCoT~\cite{mitra2023compositional}. Additionally, we benchmark CoCoT against the standard prompting baseline, which does not incorporate any CoT instructions. Note that all the experiments are conducted under the zero-shot setting. Example prompts and anwsers can be found in Fig.~\ref{fig:enter-label}. 

\noindent\textbf{Language Models.} We evaluate different prompting stratefies on two open-source LMMs: {\sc OpenFlamingo}~\cite{awadalla2023openflamingo} and MMICL~\cite{zhao2023mmicl}, as well as two proprietary models including {\sc GPT-4V}~\cite{gpt4report} and {\sc Gemini}~\cite{team2023gemini}. Due to API restrictions of {\sc GPT4-V}, we only evaluate the standard and CoCoT prompting for it. For the setting of generation, we use the default configuration for each model. We use beam search with beam width of 3 for {\sc OpenFlamingo}. In the case of {\sc MMICL}, the beam width is set to 8. For {\sc Gemini}, we opt for the API of \textit{Gemini Pro Vision} under the default settings which include a temperature of 0.4, TopK set to 32, TopP at 1, and a maximum length of 4,096. For {\sc GPT-4V}, we use the default settings of the web version as of December 30, 2023.

\subsection{Main Results}
\subsubsection{Image-to-image Matching}

\begin{figure}[h]
    \centering
    \includegraphics[width=\linewidth]{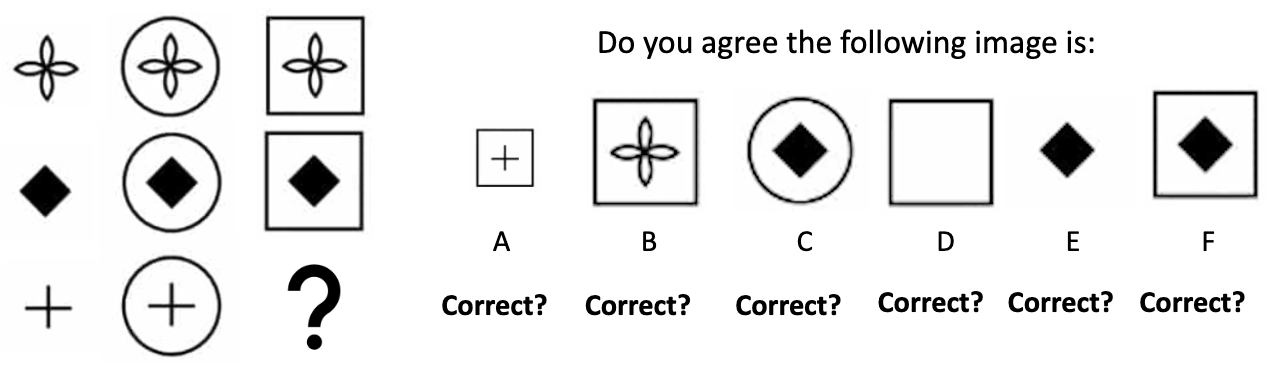}
    \vspace{-8mm}
    \caption{An example question from the image-to-image matching task, sourced from the \raven~\cite{zhang2019raven, huang2023language} dataset.}
    \label{fig:image2image_question_example}
\end{figure}

The task of image-to-image matching requires the model to extract information from two images simultaneously and then determine under a prompt whether the information from both images matches, as exemplified in Fig.~\ref{fig:image2image_question_example}. LLMs are expected to select the correct answer from the given choices. In addition to the aforementioned methods, we include another random choice baseline for comparative reference. Accuracy of LMMs with different prompting methods is shown in Table~\ref{tab:1}.

\begin{table}[t]
% \small
% \vspace{-0.1in}
\begin{center}
\caption{Accuracy of LMMs employing different prompting strategies in the image-to-image matching task. The best performance within each LMM is highlighted in \textbf{bold}.} \label{tab:1}
% \includegraphics[width=0.9\textwidth]
% OpenF indicates OpenFlamingo. * indicates the results are copied from the original paper.
\adjustbox{max width=.9\linewidth}{
\begin{tabular}{|l|c|c|}
  \hline
  % after \\: \hline or \cline{col1-col2} \cline{col3-col4} ...
    &  \raven  & \factifyv
  \\
  \hline
  Random Choice & 17.00 & 50.00 \\
  \hline
  {\sc OpenFlamingo} & 24.00 &  54.00 \\
  {\sc OpenFlamingo} + DDCoT & 24.00 & 58.40 \\ 
  {\sc OpenFlamingo} + CCoT & 24.00  &63.20 \\ 
  {\sc OpenFlamingo} + CoCoT & \textbf{26.00} & \textbf{65.00} \\ 
  \hline
  % MMICL* & 34.00  & - \\
  {\sc MMICL} & 22.00  & 64.60 \\
  {\sc MMICL} + DDCoT & 10.00  & 68.40 \\
  {\sc MMICL} + CCoT & \textbf{26.00}  & 73.20 \\
  {\sc MMICL} + CoCoT & \textbf{26.00}  & \textbf{77.00} \\
  \hline
  {\sc Gemini} & 18.00  & 58.00 \\
  {\sc Gemini} + DDCoT & 12.00 & 65.40  \\
  {\sc Gemini} + CCoT & 20.00  & \textbf{80.20} \\
  {\sc Gemini} + CoCoT & \textbf{22.00} & 77.80 \\
  \hline
  {\sc GPT-4V} & 30.00 & 74.00 \\
  {\sc GPT-4V} + CoCoT & \textbf{45.00}  & \textbf{80.60} \\
  \hline
\end{tabular}
}
\end{center}
\vspace{-0.3in}
\end{table}

\textbf{CoCoT significantly improves LMMs' performance in the image-to-image matching task.} Most models show improved performance when DDCoT and CCoT are employed, but the extent of improvement is not as significant as with CoCoT in most cases. Furthermore, regarding the {\raven} dataset, which comprises non-natural images made up of various shapes, surprisingly, {\sc Gemini} emerges as the model with the poorest performance in our evaluations when {\sc GPT-4V} performs the best which surpasses all models, including the open-source ones like {\sc OpenFlamingo} and {\sc MMICL}. 

For the {\factifyv} dataset featuring natural images, {\sc Gemini} without CoT outperforms {\sc OpenFlamingo} in similar conditions. However, when CoT is incorporated, {\sc Gemini}'s performance is almost on par with that of {\sc GPT4-V} under similar conditions. This outcome differs from the results on the {\raven} dataset, suggesting that {\sc Gemini} inherently possesses the capability to extract detailed information from natural images. Its full potential in this aspect is not fully demonstrated without the use of prompts.

In summary, our analysis of the image-to-image matching task reveals a consistent enhancement in performance across most models upon integrating various types of CoT-based prompting. This improvement underscores the ability of the visual components within LMMs to concentrate on details in terms of the task at hand. These details are subsequently processed by the LMMs for in-depth analysis, following the CoT-based prompting approach. Notably, in a majority of cases, CoCoT prompting elicits LMMs to achieve state-of-the-art performance on both natural and artificial datasets, surpassing other CoT-based strategies. This showcases the efficacy of CoCoT in guiding LMMs to accurately extract and analyze task-relevant information from images, facilitating enhanced comparative and analytical reasoning within these models.

% We initially compare the results of the state-of-the-art open-source method {\sc MMICL} and the closed-source method {\sc GPT-4V} under different prompt methods. As shown in Table~\ref{tab:1}, for {\sc GPT-4V}, due to resource limitations, we only demonstrate the comparison with and without the CoCoT prompt. It is observed that for the {\raven} datasets, {\sc GPT-4V} underperformed MMICL without prompts. However, with the addition of the CoCoT prompt, GPT-4V surpassed MMICL. This suggests that GPT-4V's image encoder can perceive detailed information but requires an appropriate prompt for information extraction and aggregation. In the case of the {\factifyv} dataset, which is closer to natural life scenarios (with more natural images), {\sc GPT-4V} already outperforms {\sc MMICL} without prompts, and the performance was further enhanced with the CoCoT prompt, also surpassing {\sc MMICL}.

\subsubsection{Multi-image-to-text Matching}

Compared to the image-to-image matching task, the multi-image-to-text matching task requires models to precisely extract information from images and match it with text. An exmaple question can be found in Fig.~\ref{fig:enter-label}. In particular, {\winoground} dataset is used for this task. Performance on {\winoground} (shown in Table~\ref{tab:2}) is assessed using three distinct metrics, each examining a different facet of the models' abilities to reason with both vision and language. The first metric, known as the \textbf{text} score, evaluates the model's capability to accurately choose the right caption when provided with an image. The second metric is the \textbf{image} score, assessing a model's ability to correctly identify the appropriate image when presented with a caption. The last metric is a composite score that integrates the first two metrics. In this \textbf{group} score, a case is considered correct if the model successfully achieves both the accurate text score and image score.

\textbf{CoCoT boosts LMMs' performance in the multi-image-to-text matching task, achieving substantial gains.} It outperforms other CoT-based methods in the majority of scenarios. This indicates that when comparing the similarities and differences of images, LMMs can better match with the text by identifying subtle differences in the input image pairs. The example in Fig.~\ref{fig:enter-label} also shows that methods like DDCoT and CCoT may miss key information, possibly as a result of misdirected focus. 

{\sc Gemini}'s performance is still the worst, indicating that although {\sc Gemini}'s visual encoder can extract detailed information from the image, the model is not able to effectively summarize the information in the image, resulting in a poor match with the text. {\sc GPT-4V}'s performance on this task is also inferior to {\sc MMICL}, indicating that {\sc GPT-4V} also struggles to effectively summarize detailed information within images, particularly when the input image pairs are very similar.
% Regarding the CoTs, CoCoT still maintains a higher standard, surpassing other CoT methods on most datasets. 

\textbf{All current models exhibit a significant gap compared to human performance.} 
The reasons for this situation could include several possibilities: 1) The visual encoder's capabilities are insufficient; 2) Large models are unable to extract detailed information from the visual encoder, thereby failing in reasoning; and 3) Large models themselves are inadequate, leading to incorrect reasoning.
Based on current experimental results, the visual encoder of LMMs is actually capable of recognizing some detailed information. However, due to the differences in the latent spaces between the visual encoder and the large language models, as well as the generalization issues of LLMs, LMMs are unable to fully understand images, resulting in mediocre performance across various tasks.

\begin{table}[t]
\begin{center}
% \small
\caption{Accuracy of LMMs employing different prompting strategies in the multi-image-to-text matching task. The best performance within each LMM is highlighted in \textbf{bold}.} \label{tab:2}
\adjustbox{max width=.9\linewidth}{
\begin{tabular}{|l|c|c|c|}
  \hline
  % after \\: \hline or \cline{col1-col2} \cline{col3-col4} ...
    &  Text & Image & Group
  \\
  \hline
  MTurk Human & 89.50 & 88.50 & 85.50 \\
  Random Choice & 25.00 & 25.00 & 16.67 \\
  % {\sc VQ2} & 47.00 & 42.00 & 30.50 \\
  % {\sc PALI} & 46.50 & 38.00 & 28.75 \\
  % {\sc Blip-2} & 44.00 & 26.00 & 23.50 \\
  \hline
  {\sc OpenFlamingo} & 39.00 & 41.25 & 33.25 \\
  {\sc OpenFlamingo} + DDCoT & 47.50 & 47.25 & 39.00 \\ 
  {\sc OpenFlamingo} + CCoT & 42.50 & 27.50 & 20.00 \\ 
  {\sc OpenFlamingo} + CoCoT & \textbf{58.25} & \textbf{55.25} & \textbf{41.50} \\ 
  \hline
  % MMICL* & 45.00 & 45.00 & 43.00 \\
  {\sc MMICL} & 46.50 & 40.75 & 37.75 \\
  {\sc MMICL} + DDCoT & 46.75 & 45.00 & 36.75 \\
  {\sc MMICL} + CCoT & 51.00 & 48.00 & 47.50 \\
  {\sc MMICL} + CoCoT & \textbf{64.25}  & \textbf{52.50} & \textbf{50.75} \\
  \hline
  {\sc Gemini} & 30.75 & 26.00 & 25.00 \\
  {\sc Gemini} + DDCoT & 45.00 & 25.00& 23.75 \\
  {\sc Gemini} + CCoT & 22.50 & \textbf{33.00} & 20.75 \\
  {\sc Gemini} + CoCoT & \textbf{40.00}  & 32.50 & \textbf{27.75} \\
  \hline
  {\sc GPT-4V} & 54.50 & 42.50 & 37.75 \\
  {\sc GPT-4V} + CoCoT & \textbf{58.50} & \textbf{49.50} & \textbf{44.50} \\
  \hline
\end{tabular}}
\end{center}
\vspace{-0.2in}
\end{table}

\subsection{Ablation Study}
CoCoT instructs LMMs to identify the similarities and differences across multiple image inputs first before providing an answer. In our ablation study, we break down the prompts into two distinct components: 1) a prompt that only requests the identification of similarities, and 2) a prompt that solely focuses on extracting the differences. As shown in Table.~\ref{tab:ablation}, we can observe that for {\sc Gemini}, the performance improves to some extent with the addition of either similarities or differences prompts alone, but not as much as when all prompts are included. For {\sc MMICL}, adding only the differences prompts leads to a minimal decrease in performance, but the best results are achieved when both prompts are incorporated.

\begin{table}[t]
% \small
% \vspace{-0.1in}
\begin{center}
\caption{Ablation study of the similarities and differences varaints of CoCoT on the {\factifyv} dataset.} \label{tab:ablation}
\adjustbox{max width=.9\linewidth}{
\begin{tabular}{|l|c|c|}
  \hline
  % after \\: \hline or \cline{col1-col2} \cline{col3-col4} ...
    & {\sc MMICL}   & {\sc Gemini}
  \\
  % \hline
  % Random Choice & 17.00 & 50.00 \\

  \hline
    No prompt & 64.60  & 58.00 \\
  + Similarities & 75.60  & 60.80 \\
  + Differences & 63.40  & 65.40 \\
  \hline
  + CoCoT & \textbf{77.00} & \textbf{77.80} \\
  \hline
\end{tabular}
}
\end{center}
\vspace{-0.2in}
\end{table}

% \vspace{-0.2in}
\section{Discussions and Conclusions}

In this study, we address the challenges faced by large multimodal models in processing detailed visual information from multiple images. We have developed the \underline{Co}ntrastive \underline{C}hain-\underline{o}f-\underline{T}hought (CoCoT) approach, a novel prompting strategy that significantly enhances LMMs' ability to discern fine-grained details in multi-image tasks. Our experiments with various models, including {\sc GPT-4V}, {\sc Gemini}, {\sc OpenFlamingo}, and {\sc MMICL}, demonstrate that CoCoT improves performance in image-to-image matching and multi-image-to-text tasks. This study contributes to the field of Artificial General Intelligence (AGI), offering new possibilities in areas requiring precise image interpretation. However, CoCoT, due to its requirement for other images for comparison, might introduce unnecessary noise. Future research should focus on refining CoCoT for more complex scenarios and integrating it with other AI technologies to further advance multimodal understanding and AGI development.

% In our exploration of Large Multimodal Models (LMMs), we tackled the challenge of processing multiple image inputs, crucial for advancing Artificial General Intelligence (AGI). Focusing on the issues of fine-grained perception and conprehending information from multiple images, we introduced the Contrastive Chain of Thought (CoCoT) approach. This novel method enhances the LMMs' ability to discern and articulate differences and similarities among various inputs, leading to significant improvements in fine-grained multi-image-to-text and image-to-image matching tasks.
% Our comprehensive evaluation, involving both open-source and closed-source models like GPT-4V, Gemini, OpenFlamingo, and MMICL, demonstrated the effectiveness of the CoCoT methodology. This approach not only overcomes the existing limitations of current models in handling complex, multi-image tasks but also propels forward the capabilities of AI systems in multimodal understanding.
% The findings of this study are pivotal, as they offer a new direction for refining the perceptual accuracy of LMMs, contributing to the broader goal of achieving more advanced AGI. The success of the CoCoT method across diverse AI models also suggests its broad applicability and potential in various AI-related applications.

\bibliographystyle{IEEEbib}
\bibliography{icme2023template}

\clearpage

% \newpage

\appendix

\section{Dataset Description}\label{appendix_sec:dataset}
\subsection{Winoground} 
The \winoground~\cite{thrush2022winoground} task involves matching images and captions which contains 400 groups of image-caption pairs. Each group contains two similar image-caption pairs. This task is challenging because the captions have the same words but in different sequences. LMMs must analyze both images and texts to identify subtle differences and understand the implied references. The {\winoground} is chosen to test if LMMs can comprehend fine-grained image information to text. Example questions are shown in Fig.~\ref{fig:dataset}. There are two tasks in the {\winoground} dataset: 1) given two images, the model is required to find out which image can match the given caption; 2) given two pieces of text, the model is required to find out which text can match the given image. 

\subsection{Raven-50}
The \raven~\cite{zhang2019raven, huang2023language} test is a common tool for assessing the nonverbal reasoning capabilities of LMMs. This test demands both visual acuity and logical reasoning to decipher the connections between images. In each scenario, participants are presented with either 3 or 8 images as inputs, alongside 6 potential answer images, each with a distinct solution. The goal is to correctly identify the appropriate image. Example questions are shown in Fig.~\ref{fig:dataset}. Note that the evaluation metric for {\sc OpenFlamingo} and {\sc MMICL} on {\raven} dataset is to calculate the logits of the output for each image pair; while for {\sc GPT-4V} and {\sc Gemini}, we directly let the model choose the correct result and calculate the accuracy.

\subsection{Factify-V}
\begin{figure*}[t]
    \centering
    \includegraphics[width=0.9\linewidth]{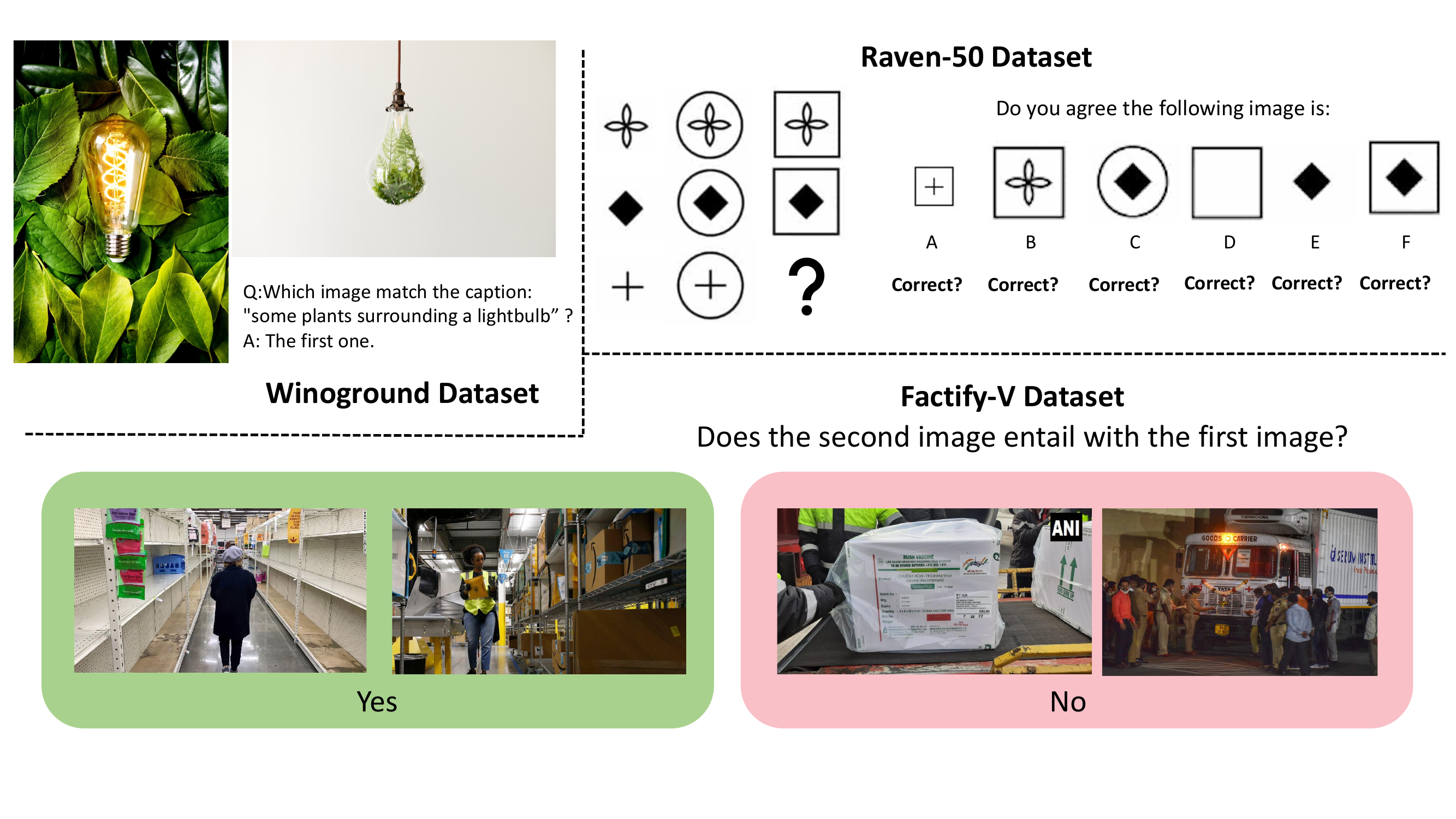}
      \vspace{-0.2in}
    \caption{Sampled questions from the {\raven}, {\factifyv}, and {\winoground} datasets.}
    \label{fig:dataset}
          \vspace{-0.1in}
\end{figure*}

The \factify~\cite{suryavardan2023factify} dataset features 35,000 data pairs for training, and 7,500 pairs each for validation and testing. Every data pair includes a claim and a corresponding document, both of which are made up of an image, text, and OCR-generated text from the image. These pairs are categorized into one of five labels: ``support multimodal'', ``support text'', ``refute'', ``insufficient multimodal'', or ``insufficient text''. Specifically, we randomly sample 500 cases in the test set, 100 for each of the 5 categories. We only use the images in the dataset in our experiments where the labels are reorganized into ``support image'' and ``refute''. The generated subset is called \factifyv. Example questions are shown in Fig.~\ref{fig:dataset}. The task involves prompting the model to determine whether the pair of input images are contextually entailed.

\section{Additional Discussions of Motivation}
As shown in Fig.~\ref{fig:enter-label}, DDCoT, leaning towards a language perspective in handling images, first decouples the original question and image information into sub-questions. It then prompts LMMs to answer these sub-questions, generating sub-answers, and finally, LMMs use these sub-questions and sub-answers to respond to the original question. CCoT, more image-oriented, initially directs LMMs to generate a Scene Graph (SG) based on image information. LMMs then use the SG's image information in conjunction with the user's question to find an answer. Given that the above methods are not effective in catching detailed information, we focus on how to enable LMMs to extract more detailed information from images, especially when the images are very similar. To address this, CoCoT is designed to guide LMMs in identifying both the similarities and the nuanced differences between images, facilitating a more in-depth and accurate interpretation of visual content.

\begin{figure*}
    \centering
    \includegraphics[width=0.9\linewidth]{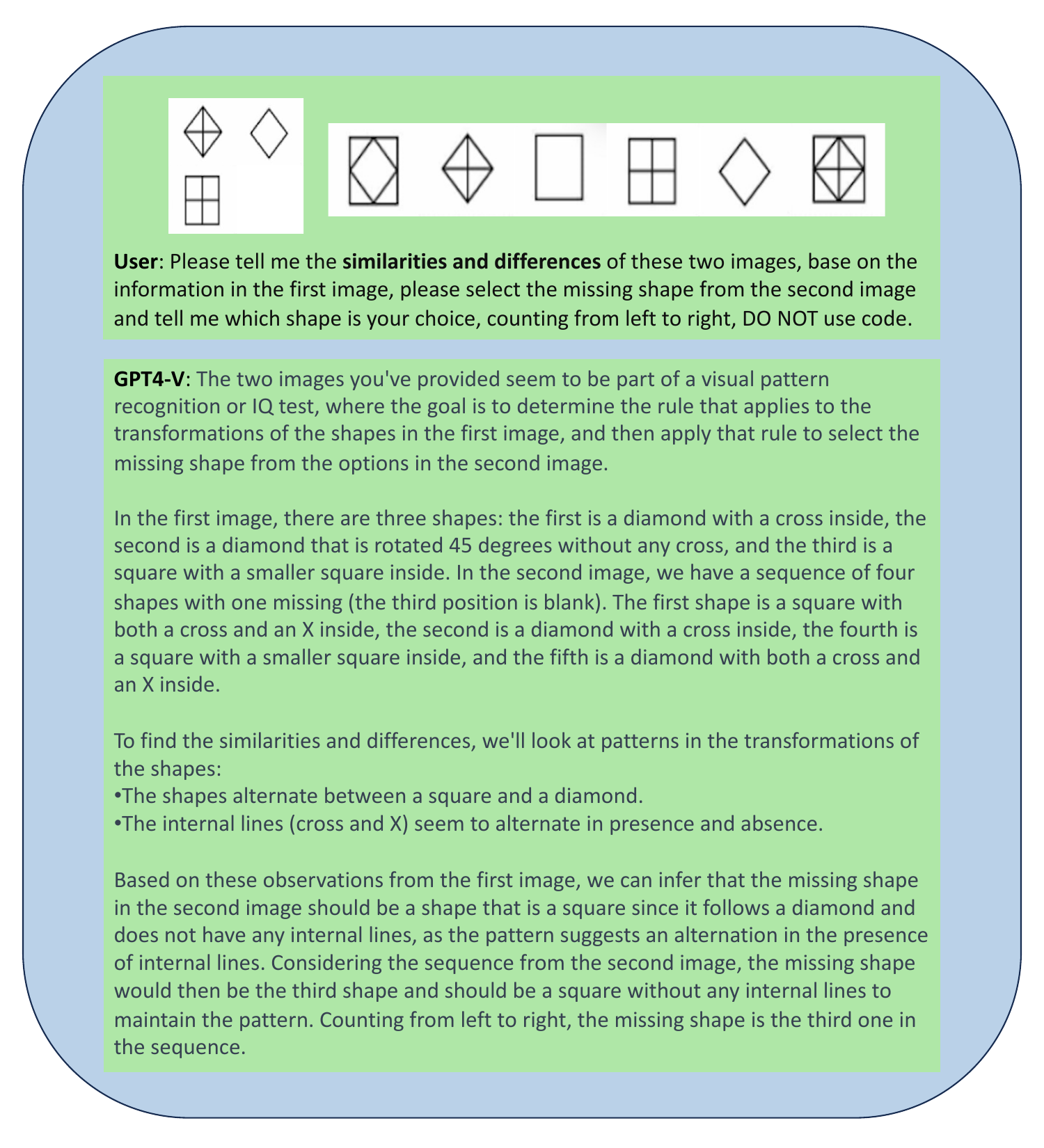}
      \vspace{-0.2in}
    \caption{An example response generated by {\sc GPT-4V} via CoCoT on the {\raven} dataset.}
    \label{fig:c1}
          \vspace{-0.1in}
\end{figure*}

% \begin{figure*}
%     \centering
%     \includegraphics[width=0.9\linewidth]{case2.pdf}
%       \vspace{-0.2in}
%     \caption{CoCoT based on GPT-4V: Incorrect Cases on the {\raven} Dataset}
%     \label{fig:c2}
%           \vspace{-0.1in}
% \end{figure*}

\begin{figure*}
    \centering
    \includegraphics[width=0.9\linewidth]{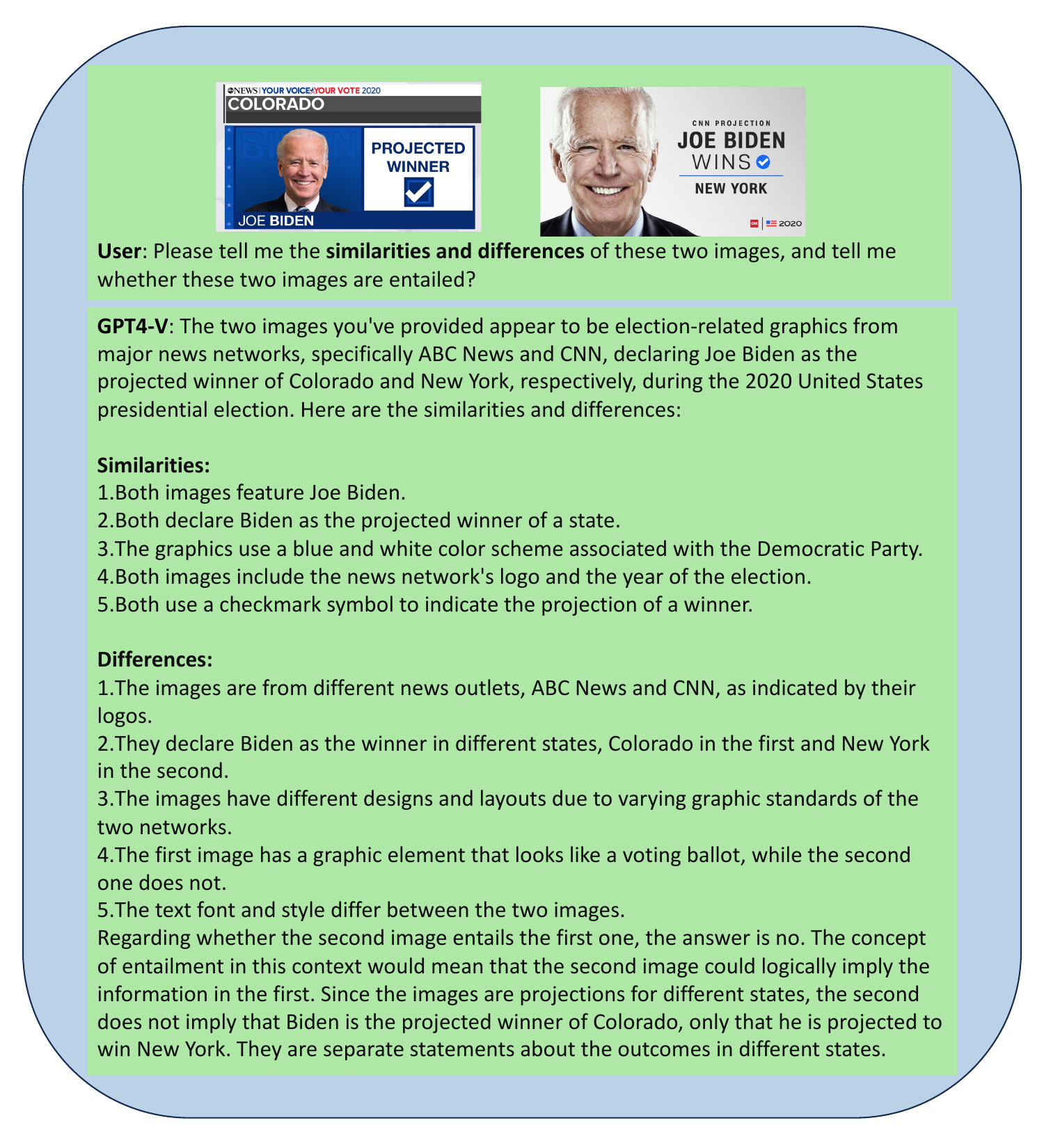}
      \vspace{-0.2in}
    \caption{An example response generated by {\sc GPT-4V} via CoCoT on the {\factifyv} dataset.}
    \label{fig:c3}
          \vspace{-0.1in}
\end{figure*}

% \begin{figure*}
%     \centering
%     \includegraphics[width=0.9\linewidth]{case4.pdf}
%       \vspace{-0.2in}
%     \caption{CoCoT based on GPT-4V: Incorrect Cases on the {\factifyv} Dataset}
%     \label{fig:c4}
%           \vspace{-0.1in}
% \end{figure*}

\section{More Examples}
We further provide  example responses of CoCoT based on {\sc GPT-4V} on different datasets, as shown in Fig.~\ref{fig:c1} and Fig.~\ref{fig:c3}.

\end{document}